\documentclass[11pt]{article}
\usepackage{acl}

\usepackage{times}
\usepackage{latexsym}

\usepackage[utf8]{inputenc}
\usepackage[T1]{fontenc}
\usepackage{amsfonts}
\usepackage{listings}
\usepackage{amssymb}
\usepackage{amsmath} 
\usepackage{algpseudocode}
\usepackage{enumitem}
\usepackage{bbm}

\usepackage{algpseudocode}  
\usepackage{mathtools} 
\usepackage{booktabs}
\usepackage[utf8]{inputenc}
\usepackage{placeins}
\usepackage{xcolor}
\usepackage{tcolorbox}
\usepackage{booktabs}      
\usepackage{colortbl}      
\usepackage{array}
\tcbuselibrary{skins,breakable}
\usepackage{multirow}
\usepackage{microtype}
\usepackage{hyperref} 
\hypersetup{
    urlcolor=black
}
\usepackage{inconsolata}

\usepackage{graphicx}
\title{Fast-Slow Thinking RM: Efficient Integration of Scalar and Generative Reward Models}

\author{
 \textbf{Jiayun Wu\textsuperscript{1}}\thanks{Work done during internship at Meituan.},
 \textbf{Peng Zhang\textsuperscript{1,}}\thanks{Corresponding authors.},
 \textbf{Yuanyuan Lu\textsuperscript{2,$^\dagger$}},
 \textbf{Shan Qu\textsuperscript{2}},
 \textbf{Ning Gu\textsuperscript{1}},
 \textbf{Tun Lu\textsuperscript{1,$^\dagger$}}
 \\[0.3em]
 \textsuperscript{1}Fudan University \quad
 \textsuperscript{2}Meituan
 \\[0.3em]
 \texttt{jiayunwu23@m.fudan.edu.cn;\ \{zhangpeng\_, ninggu, lutun\}@fudan.edu.cn}
 \\
 \texttt{\{luyuanyuan04, qushan\}@meituan.com}
}

\begin{document}
\maketitle
\begin{abstract}
Reward models (RMs) are critical for aligning Large Language Models via Reinforcement Learning from Human Feedback (RLHF). While Generative Reward Models (GRMs) achieve superior accuracy through chain-of-thought (CoT) reasoning, they incur substantial computational costs. Conversely, Scalar Reward Models (SRMs) offer efficiency but suffer from limited performance and adaptability in complex scenarios. We introduce \textbf{Fast-Slow Thinking Reward Models (F/S-RM)}, a hybrid RM architecture inspired by Dual Process Theory. It trains a single model to integrate two distinct reward paradigms: scalar-style first-token pairwise judgment (fast thinking) and CoT-based judgment (slow thinking), regulated by a dual-confidence activation mechanism that determines when to activate slow thinking. Under hybrid inference, F/S-RM achieves state-of-the-art accuracy with an average score of 84.3 across benchmarks, while reducing token consumption by 22.5\%. Code and data are publicly available\footnote{\url{https://github.com/xiaoxiaotibaiwan/F-S-RM}}.

\end{abstract}
\section{Introduction}

As Large Language Models (LLMs) continue to advance, reliably evaluating the increasingly complex and diverse text they generate has become a central challenge for defining model capability boundaries and achieving value alignment. Although human evaluation has long been regarded as the gold standard for capturing human preferences, its high cost and poor scalability significantly hinder rapid model iteration. To address these limitations, \emph{Reward Models} (RMs) have emerged as efficient proxies for approximating human preference distributions.
\begin{figure}[t]
\centering
\includegraphics[width=\columnwidth]{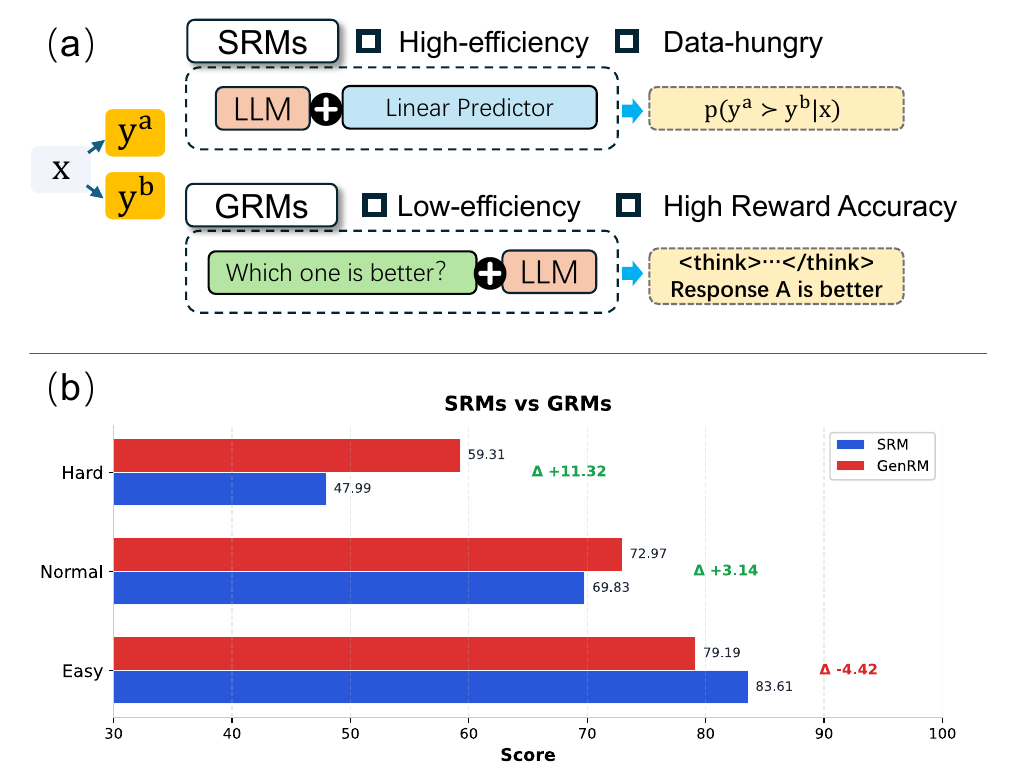}
\caption{Structural and Performance Comparison of SRMs and GRMs.
\textit{(a)} Architectural designs and output formats—SRMs produce scalar 
scores while GRMs generate reasoning chains. \textit{(b)} Performance 
across three difficulty levels on RM-Bench. GRMs excel on Hard cases while 
SRMs dominate Easy cases, revealing complementary strengths.}
\label{fig:intro}
\end{figure}
Traditional RMs predominantly adopt the form of \emph{Scalar Reward Models} (SRMs). In this paradigm, the model acts as a discriminator: a pretrained language model is augmented with a linear head that maps input text pairs to a scalar score. While SRMs offer fast inference, they collapse a complex evaluation procedure into a single-token prediction over the reward score, failing to fully exploit the generative and reasoning capabilities of LLMs. Consequently, SRMs often exhibit suboptimal accuracy, especially on out-of-distribution data \cite{sun2025rethinking}.

To address these constraints, recent research has increasingly shifted toward \emph{Generative Reward Models} (GRMs) \cite{mahan2024generative}, typically categorized under the \emph{LLM-as-a-Judge} paradigm \cite{li2025generation}. Diverging from the single scalar output of SRMs, GRMs recast the task as a generative evaluation process. By incorporating chain-of-thought (CoT) reasoning and reinforcement learning, GRMs learn to produce explicit reasoning traces before issuing final judgments, thereby mimicking human annotators' deliberate cognitive processes \cite{wang2023pandalm,zhang2024generative,chen2025rm}. This human-like "reason-then-judge" process has proven highly effective in improving evaluation accuracy, demonstrating the potential to match human expert performance \cite{chiang2023can,wang2025can}.

Despite these advantages, GRMs are not a comprehensive replacement for SRMs. Generating hundreds of reasoning tokens entails a computational cost orders of magnitude higher than the single-token calculation of SRMs, thereby limiting scalability for online use. Traditional SRMs, while benefiting from efficient inference, face their own challenges: they require massive amounts of high-quality supervision \cite{liu2025skywork,chang2025bleuberi} and often exhibit limited robustness in cross-domain generalization. 

As illustrated in Figure \ref{fig:intro}b, we analyze the performance of current open-source SRMs and GRMs across different difficulty levels on RM-Bench. While GRMs demonstrate significant advantages on complex cases, they underperform SRMs on simple tasks. This complementarity mirrors the Dual Process Theory of human cognition \cite{evans2013dual,de2023advancing}: SRMs provide efficient pattern-based 
judgments (System 1), while GRMs enable deep analytical reasoning (System 2). However, existing systems treat them as mutually exclusive alternatives rather than synergistic components. 
This gap motivates our central research question:
\textbf{Can SRMs and GRMs be efficiently integrated within the same backbone?}

In this work, we propose Fast-Slow Thinking Reward Models (\textbf{F/S-RM}), a hybrid RM architecture that integrates fast scalar judgments and slow chain-of-thought reasoning within a single base model. The model first produces a fast first-token A/B preference judgment, then employs a dual-confidence activation mechanism to determine whether to invoke slow generative reasoning with CoT for refined decisions when internal confidence is insufficient.

Experimental results demonstrate that F/S-RM achieves superior efficiency-accuracy 
trade-offs. F/S-RM achieves state-of-the-art performance with an average score of 84.3 across benchmarks while reducing token consumption by 22.5\%.
Our main contributions are:
\begin{itemize}[leftmargin=*,noitemsep,topsep=0pt]
\item We propose F/S-RM, a hybrid RM that unifies scalar and 
generative reward modeling paradigms within a backbone.
\item We design a dual-confidence activation mechanism that balances efficiency and accuracy.
\item Extensive experiments demonstrate F/S-RM's effectiveness across diverse 
benchmarks (RewardBench, RM-Bench, and JudgeBench).
\end{itemize}

\section{Preliminary}
\label{sec:preliminaries}

Reward models (RMs) play a central role in alignment pipelines such as Reinforcement Learning from Human Feedback (RLHF), converting human or AI preference signals into scalar rewards for optimization~\cite{ouyang2022training}. A RM serves as a proxy for human preference, functioning as a text classifier that evaluates the quality of a response given a specific prompt. 

\noindent\textbf{Problem Formulation.}
Formally, given an input prompt $x$ and a response $y$, the reward signal is defined as $r = R_{\psi}(x, y)$, where higher values indicate better alignment with human preferences. RMs are typically trained on preference datasets $\mathcal{D} = \{(x, y_w, y_l)\}$, where $y_w$ and $y_l$ denote the preferred and rejected responses for prompt $x$, respectively.

\noindent\textbf{Bradley-Terry Model.}
Preference learning is commonly modeled using the Bradley--Terry (BT) framework~\cite{bradley1952rank}, which assumes that observed preferences are induced by an underlying latent reward function. Under this model, the probability that $y_w$ is preferred over $y_l$ is given by 
\begin{equation}
p_{\mathrm{BT}}(y_w \succ y_l \mid x) = \sigma\big(r(x, y_w) - r(x, y_l)\big),
\end{equation}
where $\sigma(\cdot)$ denotes the sigmoid function. The reward model is trained via maximum likelihood by minimizing
\begin{equation}
\label{eq:bt_loss}
\begin{split}
\mathcal{L}_{\mathrm{BT}} = -\mathbb{E}_{(x,y_w,y_l)\sim\mathcal{D}} \Big[ \log \sigma \big( R_{\psi}(x,y_w) \\
\quad - R_{\psi}(x,y_l) \big) \Big].
\end{split}
\end{equation}

\section{Fast-Slow Thinking RM}
\begin{figure*}[t]
  \includegraphics[width=\textwidth]{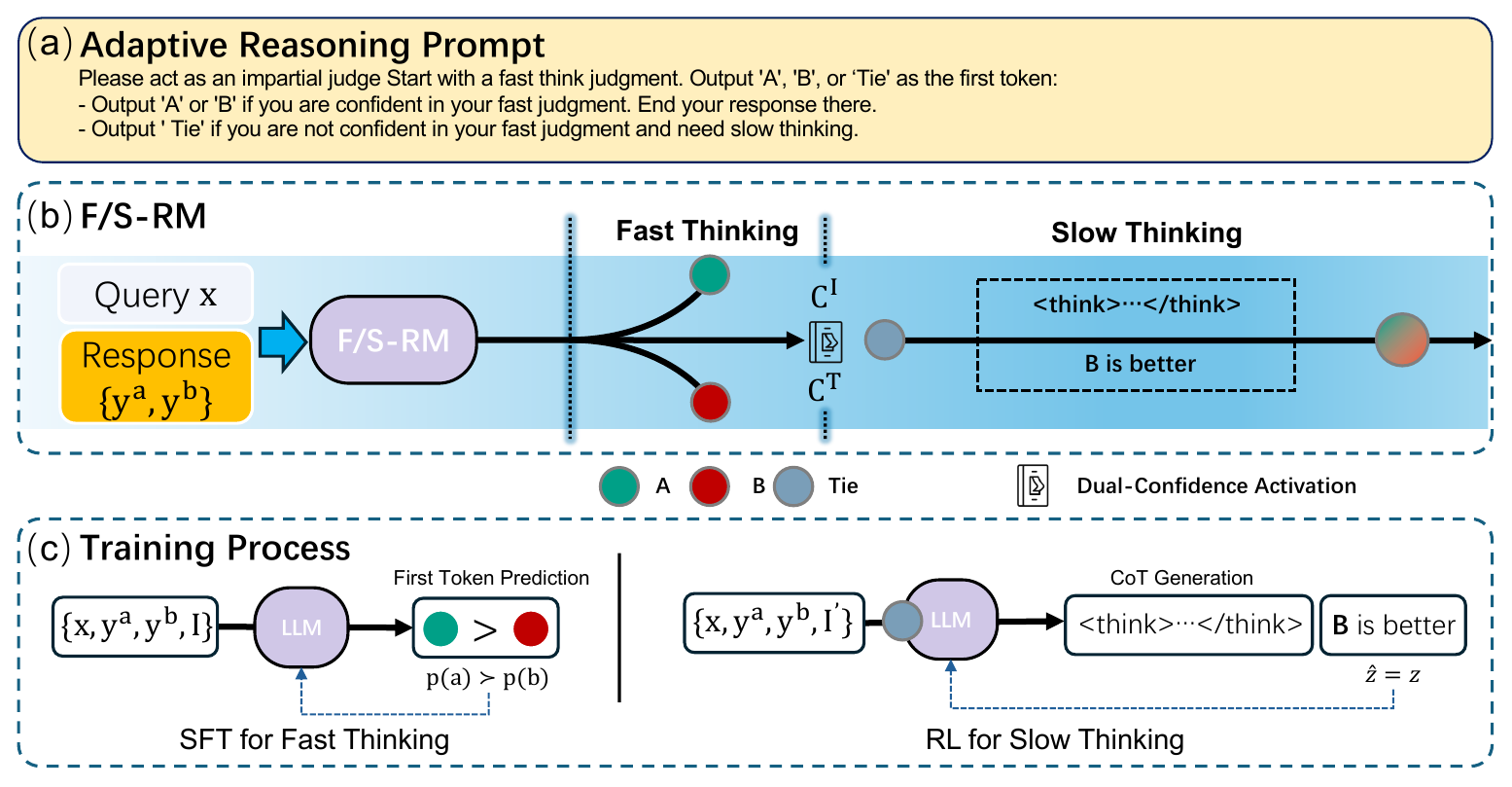}
  \caption{Architecture and training workflow of F/S-RM. \textit{(a)} Adaptive reasoning task. \textit{(b)} Reward signal generation modeled as an adaptive reasoning chain. \textit{(c)} Two-stage training pipeline for optimizing fast-thinking judgment and slow-thinking CoT reasoning.}
  \label{fig:flow}
\end{figure*}
The reward signal generation process of F/S-RM can be viewed as an adaptive reasoning chain generation process. As illustrated in Figure~\ref{fig:flow}, the model is required to make a preference judgment at the first token and activate slow thinking when its internal confidence is insufficient. However, directly training on such severely length-imbalanced samples is prone to action space collapse and reward hacking. To address this, we design a two-stage progressive training framework:

\textbf{(i) Fast Thinking as First-Token Prediction}: We model the first-token prediction as a SRM, training the model's rapid judgment capability through Bradley-Terry preference loss and token action space constraint—to obtain the fast-thinking model $\pi_\theta^{\text{fast}}$;

\textbf{(ii) RL for Slow Thinking}: Building upon $\pi_\theta^{\text{fast}}$, we append a trigger token ``tie'' to the original prompt, enabling the model to bypass fast judgment and enter slow-thinking mode with chain-of-thought reasoning. We then optimize it via online reinforcement learning (GRPO) to obtain the slow-thinking model $\pi_\theta^{\text{slow}}$;

\textbf{(iii) Dual-Confidence Activation Mechanism}: After endowing the model with both reasoning capabilities on the same foundation, we employ a dual-confidence activation strategy to determine whether to trigger slow thinking following fast judgment, yielding the final adaptive model $\pi_\theta^*$.

\subsection{Fast Thinking as First-Token Prediction}

Following \cite{zhang2024generative}, which interprets token generation probabilities as scalar rewards, we formulate \emph{fast thinking} as a first-token prediction task. This design requires only a single forward pass, matching the computational efficiency of SRMs. Under the BT framework, the training objective is to maximize the probability of the correct preference token in the first decoding step.

Formally, given a preference dataset $\mathcal{D}=\{(x,y_w,y_l)\}$, where $x$ is the user query and $(y_w,y_l)$ denotes the winning and losing responses, we construct a judge prompt $I$ that encodes both the query and response pair. The model $\pi_\theta$ predicts a preference label $z\in\{A,B\}$ at the first decoding step, where $p_\theta(z\mid I)$ denotes the probability of generating label $z$.

Training is driven by two objectives: the \emph{BT preference loss} and an \emph{action-space constraint}. Given the ground-truth preference $z^*\in\{A,B\}$, the BT loss follows the standard formulation:
\begin{equation}
\mathcal{L}_{\text{BT}}
=
-\log \sigma\!\left( \pi_\theta(z^*\mid I) - \pi_\theta(\bar{z}^*\mid I)\right),
\end{equation}
where $\bar{z}^*$ denotes the opposite label and $\sigma(\cdot)$ is the sigmoid function.

To restrict the action space to binary judgments, we impose a probability concentration constraint:
\begin{equation}
\mathcal{L}_{\text{action}}
=
-\log\!\left(\pi_\theta(A\mid I)+\pi_\theta(B\mid I)\right),
\end{equation}
which penalizes probability leakage to irrelevant tokens, ensuring that the model concentrates its output distribution on the two preference labels.

The overall supervised fine-tuning objective is
\begin{equation}
\mathcal{L}_{\text{SFT}}
=
\mathcal{L}_{\text{BT}}+\lambda \mathcal{L}_{\text{action}},
\end{equation}
where $\lambda>0$ controls the strength of the action-space constraint. Optimizing this objective yields the fast-thinking model $\pi_\theta^{\text{fast}}$.
\subsection{RL for Slow Thinking}

Building upon the fast-thinking model $\pi_\theta^{\text{fast}}$, we employ GRPO reinforcement learning~\cite{guo2025deepseek} to activate its complex reasoning capability. The key design is a \emph{trigger token mechanism}: by appending a special token ``tie'' to the judge prompt $I$, we construct an augmented prompt $I' = I \oplus \text{``tie''}$ that instructs the model to bypass fast judgment and enter slow-thinking mode. This lightweight switch incurs only one additional token while enabling step-by-step reasoning for ambiguous samples.

We optimize the slow-thinking policy via GRPO, using $\pi_\theta^{\text{fast}}$ as the reference policy $\pi_{\text{ref}}$. The training objective combines a clipped policy gradient with KL regularization:
\begin{equation}
\begin{split}
\mathcal{L}_{\text{GRPO}} 
&= \mathbb{E}_{\tau \sim \pi_\theta} \bigg[ \min\Big( \rho(\tau)A(\tau), \, \\
&\qquad \text{clip}\big( \rho(\tau), 1-\epsilon, 1+\epsilon \big)A(\tau) \Big) \bigg] \\
&\quad - \beta D_{\text{KL}}(\pi_\theta \| \pi_{\text{ref}}),
\end{split}
\end{equation}
where $\rho(\tau)=\pi_\theta(\tau)/\pi_{\text{ref}}(\tau)$. Following a group-based, critic-free design, the advantage is computed by sampling $K$ trajectories per prompt and normalizing terminal rewards within each group,
$
A(\tau)=\frac{r(\tau)-\mu_{\text{prompt}}}{\sigma_{\text{prompt}}}.$
The slow-thinking policy shares the same backbone as the fast-thinking model, enabling stable continuation training.
Reward design combines structural and semantic supervision through a single reward:
\begin{equation}
R(\tau)=
\begin{cases}
1, & \text{if the format is correct and } \hat z=z,\\
0, & \text{if the format is correct and } \hat z\neq z,\\
-1, & \text{if the format is incorrect},
\end{cases}
\end{equation}
where $z$ is the ground-truth label and $\hat z$ is the predicted decision. This design first penalizes format violations and then rewards correct predictions under the prescribed two-stage generation format. Optimizing $\mathcal{L}_{\text{GRPO}}$ yields the slow-thinking model $\pi_\theta^{\text{slow}}$.

\subsection{Dual-Confidence Activation with Prefix Reuse}

F/S-RM supports two modes on a shared backbone: $\pi_\theta^{\text{fast}}$ for direct judgment and $\pi_\theta^{\text{slow}}$ for CoT reasoning. These two modes share the same judge prompt, and differ only by the appended trigger token ``tie''. This one-token difference enables maximal prefix reuse: all inputs first undergo fast thinking, and slow thinking is activated only by appending ``tie'' when the fast decision is uncertain. Therefore, the key challenge is determining when to activate the more expensive slow-thinking process. We address this via a \emph{dual-threshold activation mechanism} based on two complementary confidence metrics.



\textbf{Intuition Confidence.} The first metric, $C^I$, measures the discriminative certainty between the two candidate labels:
\begin{equation}
C^I = \big|\pi_\theta^{\text{fast}}(A \mid I) - \pi_\theta^{\text{fast}}(B \mid I)\big|.
\end{equation}
A low $C^I$ indicates ambiguity between options, suggesting the need for deeper reasoning.

\textbf{Token Confidence.} The second metric, $C^T$, quantifies distribution sharpness by measuring probability leakage to irrelevant tokens. Inspired by recent findings that token distributions reflect reasoning quality~\cite{kang2025scalable,fu2025deep}, we define:
\begin{equation}
C^T = \frac{1}{k} \sum_{x \in \text{TopK}(\mathcal{V} \setminus \mathcal{S})} -\log \pi_\theta^{\text{fast}}(x \mid I),
\end{equation}
where $\mathcal{V}$ is the vocabulary, $\mathcal{S}=\{A, B\}$ is the candidate set, and TopK selects the $k$ most probable tokens outside $\mathcal{S}$. A lower $C^T$ indicates significant probability mass leaking to irrelevant tokens, signaling high uncertainty.

\textbf{Activation Rule.} Slow thinking is triggered when both confidence metrics fall below their respective confidence thresholds:
\begin{equation}
C^I < \tau_I \quad \text{and} \quad C^T < \tau_T,
\end{equation}
where $\tau_I$ and $\tau_T$ are the intuition and token confidence thresholds, respectively, set to the mean values of $C^I$ and $C^T$ computed on correctly-predicted samples in the training set. This dual-threshold design ensures that slow thinking is activated only when the model exhibits both low discriminative confidence and high distributional uncertainty, yielding the final Fast/Slow model $\pi_\theta^*$.

\section{Experiments}
\begin{table*}[!t]
\centering
\small
\begin{tabular}{lcccc}
\toprule
\textbf{Model} & \textbf{RewardBench} & \textbf{RM-Bench} & \textbf{JudgeBench} & \textbf{Avg.} \\
\midrule
\multicolumn{5}{l}{Scalar Reward Models} \\
\midrule
Llama-3-OffsetBias-RM-8B & 89.0 & 71.3 & 63.5 & 74.6 \\
ArmoRM-Llama3-8B-v0.1 & 90.4 & 69.3 & 59.7 & 73.1 \\
Internlm2-20b-reward & 90.2 & 68.3 & 64.3 & 74.3 \\
Skywork-Reward-Llama-3.1-8B-v0.2 & 93.1 & 72.1 & 62.9 & 76.0 \\
LDL-Reward-Gemma-2-27B-v0.1 & \underline{95.0} & 71.1 & 64.2 & 76.8 \\
Skywork-Reward-Gemma-2-27B-v0.2 & 94.3 & 70.0 & 66.5 & 76.9 \\
Llama-3.1-Nemotron-70B & 93.9 & 72.2 & 65.8 & 77.3 \\
INF-ORM-Llama3.1-70B & \textbf{95.1} & 73.8 & 70.2 & 79.7 \\
Skywork-Reward-V2-Qwen3-4B\cite{liu2025skywork} & 93.4 & 81.6 & 69.3 & 81.4 \\
Skywork-Reward-V2-Qwen3-8B & 93.7 & 82.6 & \textbf{73.4} & \underline{83.2} \\
\midrule
\multicolumn{5}{l}{LLM-as-a-Judge \& Generative Reward Models} \\
\midrule
GPT-4o-0806 & 86.7 & 72.5 & 59.8 & 73.0 \\
Claude-3.5-Sonnet & 84.2 & 61.0 & 70.0 &  \\
Gemini-1.5-pro & 88.2 & 75.2 &  56.5 & 73.3 \\
DeepSeek-GRM-27B & 88.5 & 72.7 & 69.0 & 76.7 \\
RM-R1-Qwen-Instruct-32B & 91.4 & 79.1 & -- & -- \\
RM-R1-DeepSeek-Distill-Qwen-32B & 90.9 & 83.9 & -- & -- \\
EvalPlanner-Llama-3.1-70B & 93.9 & 80.0 & 50.9 & 74.9 \\
EvalPlanner-Llama-3.3-70B & 93.8 & 82.1 & 56.6 & 77.5 \\
J1-Llama-8B & 85.7 & 73.4 & 42.0 & 67.0 \\
J1-Llama-70B & 93.3 & 82.7 & 60.0 & 78.7 \\
BR-RM-Qwen-8B\cite{jiao2025think} & 91.0 & 85.0 & -- & -- \\
BR-RM-Qwen-14B & 92.1 & 85.9 & -- & -- \\
Uncertainty-Router-8B\cite{xu2026ask} & 91.6 & 81.9 & -- & -- \\ 
\midrule
\multicolumn{5}{l}{Our F/S-RM} \\
\midrule
F/S-RM-Qwen3-4B-Hybrid& 90.5& 84.5 & 69.3 & 81.4   \\
\rowcolor{cyan!15}
\quad $\Delta$ vs. Slow & -0.05 & -0.11 & -0.14 & -0.10 \\
\rowcolor{cyan!15}
\quad $\downarrow$ Saving Tokens  & \textcolor{green!70!black}{31.2\%} & \textcolor{green!70!black}{19.9\%} & \textcolor{green!70!black}{6.5\%} & \textcolor{green!70!black}{19.2\%} \\
F/S-RM-Qwen3-8B-Hybrid& 92.0& \underline{87.9}& \underline{73.0} &  \textbf{84.3}  \\
\rowcolor{cyan!15}
\quad $\Delta$ vs. Slow & +0.08 & -0.34 & +0.57  & +0.10 \\
\rowcolor{cyan!15}
\quad $\downarrow$ Saving Tokens  & \textcolor{green!70!black}{39.1\%} & \textcolor{green!70!black}{22.3\%} & \textcolor{green!70!black}{6.2\%} & \textcolor{green!70!black}{22.5\%} \\
\midrule
\multicolumn{5}{l}{Our F/S Only Models} \\
\midrule
F/S-RM-Qwen3-4B-Fast& 85.9 & 77.7& 67.0 & 76.9  \\
F/S-RM-Qwen3-4B-Slow& 90.5 & 84.6& 69.4  & 81.5   \\
\midrule
F/S-RM-Qwen3-8B-Fast& 87.5& 80.3& 65.1 & 77.6   \\
F/S-RM-Qwen3-8B-Slow& 91.9& \textbf{88.2}& 72.4 & \underline{84.2}   \\ 
\hline
\end{tabular}
\caption{Comparison on RewardBench, RM-Bench, JudgeBench, and average performance. Bold numbers indicate the best performance, Underlined numbers indicate the second best. $\Delta$ shows hybrid performance change vs. full slow thinking; $\downarrow$ shows token reduction vs. full slow thinking. Detailed comparison results are provided in the Appendix \ref{sec:Domain-Specific Comparison}.}
\label{tab:model_comparison}
\end{table*}
\textbf{Training Setup.} Following \cite{chen2025rm}, we randomly sample 20K instances from the dataset as our training set to align with other methods in terms of data scale. We select Qwen3-4B/8B as our base models, which possess inherent reasoning capabilities that facilitate the activation of slow thinking. For the Reinforcement Learning phase, we utilize the VERL framework~\cite{sheng2024hybridflow} on 4 NVIDIA A100-80G GPUs. Implementation details and hyperparameter settings are provided in Appendix \ref{sec:Training Dataset}.

\textbf{Evaluation Datasets.} We evaluate our method on three main benchmarks: (1) \textbf{RewardBench}~\cite{lambert2025rewardbench}, the most widely used reward model benchmark covering chat, safety, reasoning, and prior sets, though it has known biases~\cite{wu2025rewordbench}; (2) \textbf{RM-Bench}~\cite{liu2024rm}, designed to address RewardBench's limitations with stronger focus on sensitivity to subtle content differences, robustness against stylistic biases, and correlation with policy model performance; (3) \textbf{JudgeBench}~\cite{tan2024judgebench}, a challenging benchmark targeting objective judgment in complex tasks. Detailed descriptions of these datasets are provided in the Appendix \ref{sec:Evaluation Datasets}.

\subsection{Main Results}

Table~\ref{tab:model_comparison} compares F/S-RMs with strong baselines across three benchmarks. Under hybrid inference, F/S-RM-Qwen3-8B achieves an average score of 84.3 across benchmarks while reducing token consumption by 22.5\%. The token-reduction metric is described in Appendix~\ref{app: Metric}.


\noindent \textbf{Baseline Analysis}
~Overall, SRMs perform well on relatively simple tasks such as RewardBench, but underperform on reasoning-intensive benchmarks, where GRMs, especially ReasonRMs with chain-of-thought reasoning, show clear advantages. Our fast and slow models exhibit the same trend. We further compare with Uncertainty-Router-8B~\cite{xu2026ask}, a routing-based baseline that uses uncertainty from a Llama-3.1-8B-Instruct reward model to decide whether to query DeepSeek-R1; we report its best operating point, while our method uses no external judge. Notably, two strong baselines share the same model backbone: \underline{Skywork-Reward-V2}~\cite{liu2025skywork} and the \underline{BR-RM-Qwen}~\cite{jiao2025think} series, both trained on Qwen3 architectures. Skywork-Reward-V2-Qwen3 is an exemplary SRM trained on 20M human-enhanced annotations, while BR-RM-Qwen represents state-of-the-art reason-then-judge GRMs with multi-round branching reasoning.

\noindent\textbf{F/S-RM Achieves Outstanding Performance with Lower Token Consumption.}
~Our two-stage training effectively enhances slow-thinking capability: F/S-RM-Qwen3-8B-Slow achieves the best RM-Bench score of 88.2 among comparable models. More importantly, F/S-RM-Qwen3-8B-Hybrid maintains near-identical performance while reducing token computation by 22.5\% on average, and even improves over pure slow modes on JudgeBench by up to 0.57. Across benchmarks, the hybrid model reaches 92.0 on RewardBench, 87.9 on RM-Bench, and 73.0 on JudgeBench, outperforming strong Qwen3-based baselines such as BR-RM-Qwen-8B on both performance and efficiency. These results suggest that Fast and Slow thinking are complementary: clear cases can be resolved efficiently through Fast judgment, while uncertain cases benefit from deliberate Slow reasoning.

\subsection{Ablation Study}


\begin{table}[t]
\centering
\small
\renewcommand{\arraystretch}{1.1}
\begin{tabular}{@{}lccc@{}}
\toprule
\textbf{Configuration} & \textbf{RB} & \textbf{RM-B} & \textbf{JB} \\
\midrule
\multicolumn{4}{@{}l@{}}{\textit{Fast-Thinking Performance (4B)}} \\
Fast (SFT only) & 86.0 & 77.8 & 66.9 \\
Fast (SFT+RL) & 85.9 & 77.7 & 67.0 \\
\rowcolor{cyan!15}
\quad $\Delta$ & -0.1 & -0.1 & +0.1 \\
\midrule
\multicolumn{4}{@{}l@{}}{\textit{Fast-Thinking Performance (8B)}} \\
Fast (SFT only) & 87.8 & 80.1 & 64.4 \\
Fast (SFT+RL) & 87.5 & 80.2 & 65.1 \\
\rowcolor{cyan!15}
\quad $\Delta$ & -0.3 & +0.1 & +0.7 \\
\midrule
\multicolumn{4}{@{}l@{}}{\textit{Slow-Thinking Performance (4B)}} \\
Slow (RL only) & 89.8 & 85.0 & 55.1 \\
Slow (SFT+RL) & 90.5& 84.5 & 69.4 \\
\rowcolor{cyan!15}
\quad $\Delta$ & +0.7 & -0.5 & +14.3 \\
\midrule
\multicolumn{4}{@{}l@{}}{\textit{Slow-Thinking Performance (8B)}} \\
Slow (RL only) & 87.6 & 83.4 & 70.7 \\
Slow (SFT+RL) & 91.9 & 88.2 & 72.4 \\
\rowcolor{cyan!15}
\quad $\Delta$ & +4.3 & +4.8 & +1.7 \\
\bottomrule
\end{tabular}
\caption{Ablation study on fast-slow training. RB, RM-B, and JB denote RewardBench, RM-Bench, and JudgeBench, respectively. \textbf{Fast (SFT only)}: fast-thinking performance after SFT training only. \textbf{Fast (SFT+RL)}: fast-thinking performance after both SFT and RL stages. \textbf{Slow (RL only)}: slow-thinking performance with RL training only (skipping SFT). \textbf{Slow (SFT+RL)}: slow-thinking performance after complete two-stage training.}
\label{tab:ablation}
\end{table}
Table~\ref{tab:ablation} presents the results across two dimensions: (1) the impact of RL on fast-thinking performance, and (2) the necessity of SFT for slow-thinking training.

\textbf{RL Preserves Fast-Thinking Capability.} Comparing Fast (SFT only) and Fast (SFT+RL), we observe minimal degradation after RL training. For the 8B model, changes are negligible across benchmarks ($\Delta \leq 0.7$) and even slightly positive on JudgeBench (+0.7), indicating that slow-thinking RL does not interfere with fast-thinking capability.
\textbf{Fast-Thinking Training Facilitates Slow Thinking.}
Comparing Slow (RL only) and Slow (SFT+RL), we find that first-token SFT improves both slow-thinking performance and stability, especially on challenging benchmarks. For the 8B model, SFT yields clear gains of +4.3 on RewardBench, +4.8 on RM-Bench, and +1.7 on JudgeBench. The effect is stronger for stability in the 4B model: Slow (RL only) collapses on JudgeBench (55.1), often failing to produce a definitive verdict, while SFT+RL restores robustness to 69.4. This suggests that the SFT stage provides intuitive judgment foundations that support more stable slow reasoning.


\subsection{Efficient Hybrid Inference with Prefix Reuse}

F/S-RM is not a two-model cascade: Fast and Slow thinking share the same backbone and the same judge-prompt prefix. At inference time, the model first obtains the first-token probabilities for Fast judgment. High-confidence samples terminate immediately, while uncertain samples append the trigger token \texttt{tie} and continue Slow reasoning from the shared prefix states. Fast-only judgment requires only one output token and incurs negligible latency compared with full Slow reasoning: on Qwen3-4B/8B, Fast inference takes only 46.52/117.47 ms, compared with 21.48/27.03 s for Slow inference. This large cost gap provides the basis for efficient selective activation.

To verify that this advantage translates into end-to-end speedups, we measure the wall-clock latency of Qwen3-8B Hybrid inference on one NVIDIA A100 with batch size 1. As shown in Table~\ref{tab:hybrid_latency}, Hybrid inference consistently reduces latency relative to full Slow inference across all three benchmarks, with an average reduction of 27.1\%. This is consistent with the reduction in generated tokens and confirms that selectively terminating confident cases through the Fast path yields practical inference savings.

\begin{table}[h]
\centering
\small
\setlength{\tabcolsep}{5pt}
\begin{tabular}{@{}lrrr@{}}
\toprule
\textbf{Dataset} &
\textbf{Hybrid} &
\textbf{Full Slow} &
\textbf{Saving} \\
\midrule 
RewardBench & 11.855 & 21.186 & 44.0\% \\
RM-Bench    & 16.280 & 25.139 & 35.2\% \\
JudgeBench  & 29.044 & 32.093 & 9.5\% \\
\midrule
Average     & \textbf{19.060} & 26.139 & \textbf{27.1\%} \\
\bottomrule
\end{tabular}
\caption{End-to-end wall-clock latency (seconds per case) of F/S-RM-Qwen3-8B Hybrid and full Slow inference.}
\label{tab:hybrid_latency}
\end{table}

The shared-prefix design further reduces the overhead of entering the Slow path. Since Slow reasoning differs only by the appended trigger token \texttt{tie}, the existing prefix states can be directly reused rather than recomputed. This optimization mainly reduces redundant prefill computation during Fast-to-Slow switching, while the overall latency reduction is primarily contributed by Fast exits. Detailed prefix-reuse analysis is provided in Appendix~\ref{app:Prefix Reuse}.
\begin{table*}[th]
\centering
\small
\begin{tabular}{lrrrrr}
\toprule
\textbf{Dataset} & \textbf{Both Correct} & \textbf{Slow Fixes} & \textbf{Fast Corrects} & \textbf{Both Wrong} & \textbf{Oracle Ceiling} \\
\midrule
RewardBench & 4830 (80.9\%) & 671 (11.2\%) & 402 (6.7\%) & 67 (1.1\%) & 98.9\% \\
RM-Bench & 6056 (76.1\%) & 952 (12.0\%) & 331 (4.2\%) & 623 (7.8\%) & 92.2\% \\
JudgeBench & 325 (46.4\%) & 180 (25.7\%) & 127 (18.1\%) & 68 (9.7\%) & 90.3\% \\
\bottomrule
\end{tabular}
\caption{Complementarity between fast-only and slow-only decisions on Qwen3-8B. ``Slow Fixes'' denotes fast-wrong/slow-correct cases; ``Fast Corrects'' denotes fast-correct/slow-wrong cases. Oracle ceiling is the accuracy achieved by choosing the correct mode whenever either mode is correct.}
\label{tab:complementarity}
\end{table*}



\subsection{Efficacy of Dual-Confidence Activation}

\begin{figure}[h]
\includegraphics[width=\columnwidth]{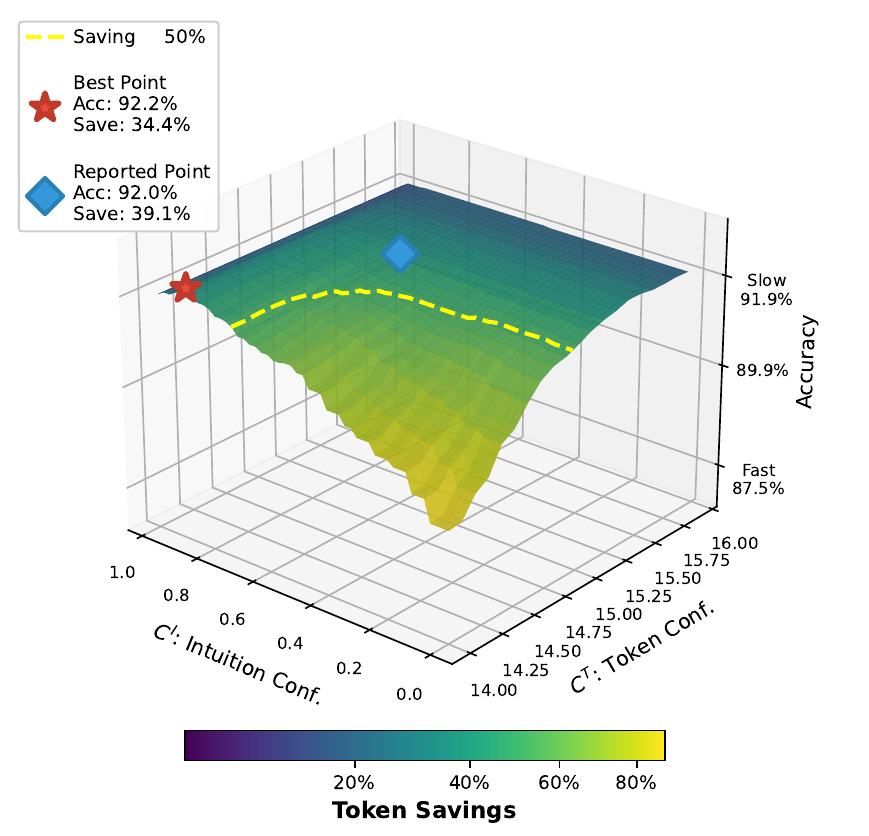}
\caption{Dual-Confidence visualization on RewardBench (Qwen3-8B). Vertical axis ($Z$): hybrid accuracy; surface color: token savings percentage; horizontal axes: $C^I$ (left) and $C^T$ (right). Deeper yellow indicates higher savings; height indicates better performance.}
\label{fig:3d}
\end{figure}

\begin{figure}[h]
\includegraphics[width=\columnwidth]{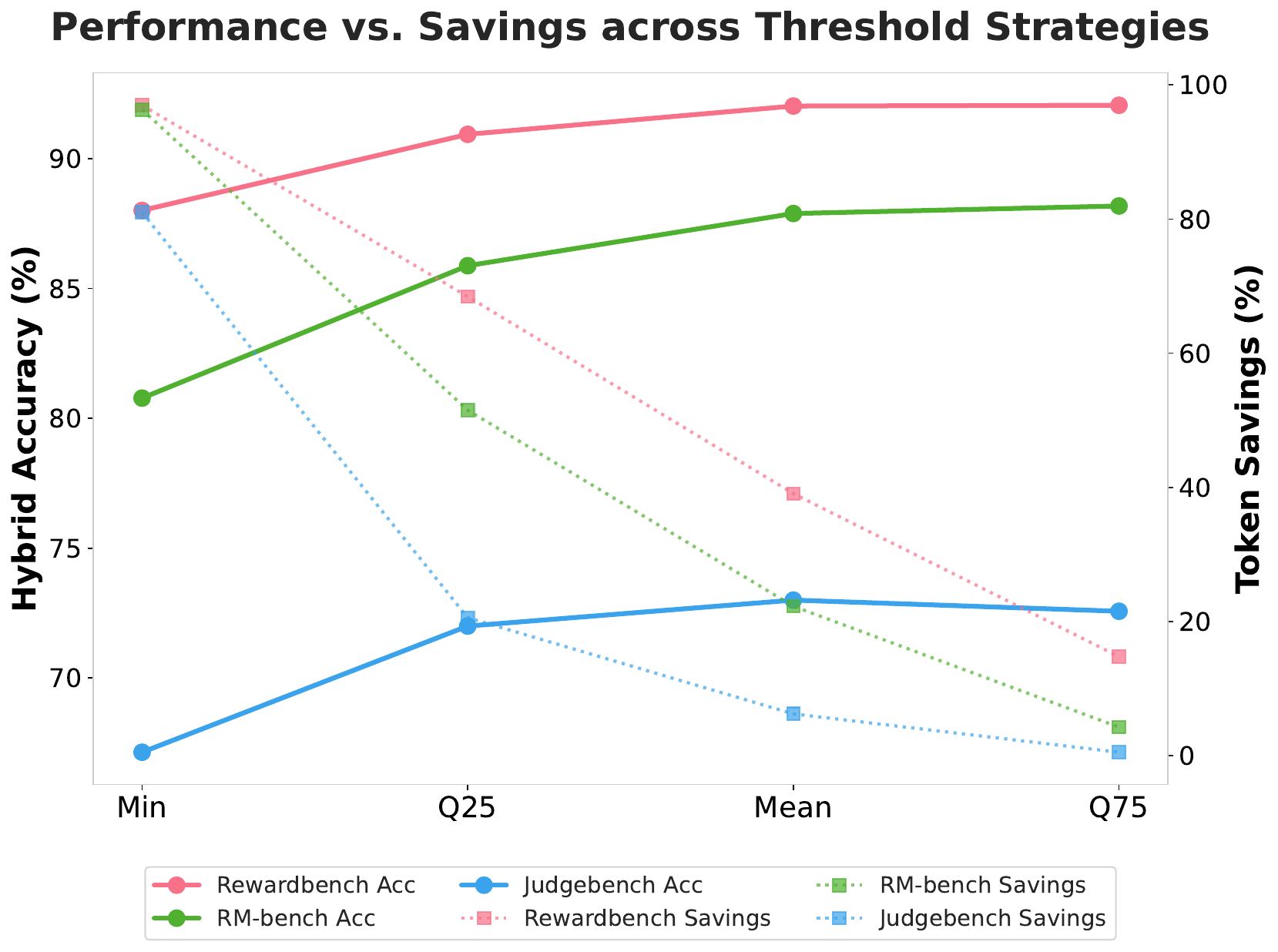}
\caption{Performance-efficiency trade-off across threshold settings. Thresholds are set at minimum, 25th percentile, mean (main results), and 75th percentile of $C^I$ and $C^T$ from correctly predicted training samples. Left y-axis: hybrid accuracy; right y-axis: token savings.}
\label{fig:threshold}
\end{figure}

\begin{figure}[h]
\includegraphics[width=\columnwidth]{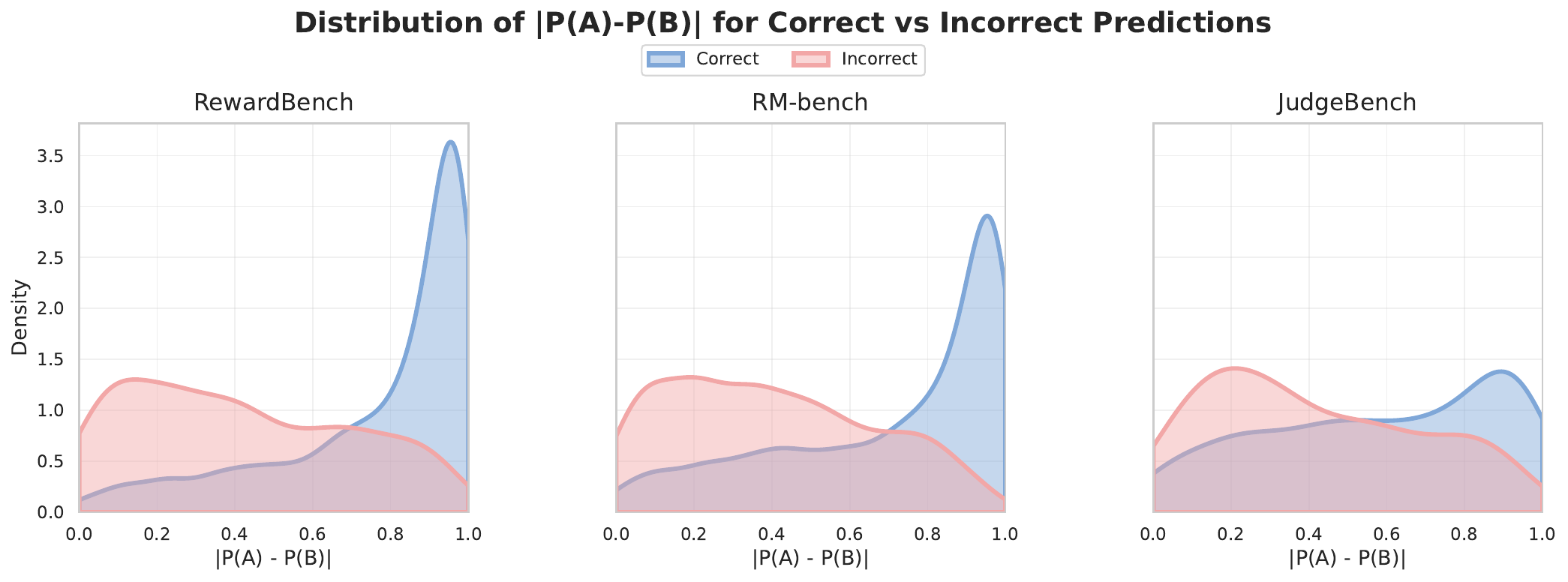}
\caption{Distribution of intuition Confidence $C^I$ for correct vs. incorrect predictions across RewardBench, RM-Bench, and JudgeBench; correct predictions concentrate at higher margins, indicating a positive but imperfect confidence-accuracy alignment.}
\label{fig:confidence_distribution}
\end{figure}

\textbf{Stability of Dual Confidence.}
Figure~\ref{fig:3d} demonstrates the performance-efficiency trade-off of dual Confidence across a wide range of configurations. Taking Qwen3-8B on RewardBench as an example, the plot reveals a distinct ``efficiency plateau'' region where the hybrid model maintains near-optimal performance while achieving substantial token reduction. The optimal point identified through grid search differs only slightly from our reported configuration, demonstrating the robustness of dual-confidence activation.

\textbf{More relaxed threshold settings enable even greater token savings.}
Our dual-confidence mechanism represents a performance-oriented approach that jointly considers both intuition confidence ($C^I$) and token confidence ($C^T$) to determine when to activate slow thinking. In Appendix~\ref{app:dual_confidence_ablation}, we provide a comprehensive comparison between dual-confidence and single-confidence routing strategies. As shown in Figure~\ref{fig:threshold}, across all three benchmarks, configurations between the Q25 and Mean thresholds achieve substantial token savings with only small performance changes. The same model-specific thresholds are applied across RewardBench, RM-Bench, and JudgeBench without benchmark-specific tuning.

\textbf{Confidence Calibration of Fast Thinking.}
Figure~\ref{fig:confidence_distribution} shows the distribution of intuition confidence $C^I=|P(A)-P(B)|$ for correct and incorrect Fast predictions. Correct predictions are concentrated in high-confidence regions, whereas incorrect predictions are skewed toward low-confidence regions. This separation supports the use of Fast-thinking confidence as a routing signal for activating Slow reasoning.

\begin{table}[t]
\centering
\small
\setlength{\tabcolsep}{6pt}
\renewcommand{\arraystretch}{1.08}
\begin{tabular}{@{}lrr@{}}
\toprule
\textbf{Routing outcome} & \textbf{Count} & \textbf{Ratio} \\
\midrule
Direct exit & 5,687 & 38.9\% \\
\quad Fast errors & 214 & 1.46\% \\
\quad Slow-fixable errors missed & 64 & 0.44\% \\
\addlinespace[2pt]
Fast errors corrected by Slow & 1,459 & 10.0\% \\
\midrule
\multicolumn{2}{l}{\textbf{Slow-fixable error capture rate}} 
& \textbf{95.8\%} \\
\bottomrule
\end{tabular}
\caption{Reliability of dual-confidence routing on Qwen3-8B across the three benchmarks.}
\label{tab:routing_reliability}
\end{table}

\textbf{Routing Reliability.}
Confidence-based routing cannot identify every Fast error, particularly when the model is confidently wrong. Across 14,632 evaluated samples, 1,523 Fast errors are correctable by Slow reasoning, of which the router successfully escalates 1,459, corresponding to a 95.8\% capture rate. Only 64 Slow-fixable Fast errors are missed by routing, accounting for 0.44\% of all evaluated samples. A more conservative routing configuration further reduces these missed escalations at the cost of fewer Fast exits. The complete error breakdown and threshold-reliability analysis are provided in Appendix~\ref{app:routing_reliability}.

\subsection{Fast-Slow Complementarity Analysis.}

Table~\ref{tab:complementarity} shows that fast and slow thinking are complementary rather than redundant. Slow reasoning corrects many fast-mode mistakes, while fast judgment also succeeds in cases where slow reasoning fails. The resulting oracle ceilings are substantially higher than current hybrid performance, suggesting that the main remaining opportunity lies in better routing. The sizeable ``Fast Corrects'' population shows that slow thinking is not uniformly superior and may overthink some cases, while ``Slow Fixes'' confirms the value of deliberate reasoning on ambiguous examples. This explains why adaptive routing can outperform slow-only inference on JudgeBench and motivates richer uncertainty signals for mode selection.

\subsection{Domain-Specific Analysis}
\textbf{Difficulty- and Domain-Aware Routing.}
The fine-grained results in Appendix~\ref{sec:Domain-Specific Comparison} show that F/S-RM exhibits difficulty-aware routing behavior, assigning easier and clearer cases more often to fast thinking while reserving slow thinking for more complex domains.

\section{Related Work}
\subsection{Scalar Reward Models}
Reward models (RMs) are central to alignment techniques like RLHF, converting preference signals into scalar rewards~\cite{ouyang2022training}. SRMs are typically trained on preference datasets using the Bradley--Terry framework~\cite{bradley1952rank}, where they learn to predict which response is preferred by modeling preference probabilities through sigmoid functions~\cite{liu2024skywork}. SRMs augment pretrained language models with a regression head to map representations to scalar scores. This simple, efficient approach makes SRMs popular in alignment pipelines, but their reliance on large-scale preference data and inability to handle complex multi-dimensional evaluations can limit generalization across domains~\cite{liu2025skywork}.
\subsection{Generative Reward Models with Reasoning}
GRMs~\cite{mahan2024generative}, including Reasoning Reward Models (RRMs)~\cite{guo2025reward,jiao2025think}, leverage LLMs' generative capacity to produce explicit judgments and rationales, enabling inference-time reasoning such as long CoT. Closely related to LLM-as-a-Judge~\cite{li2024llms}, GRMs approximate human decision-making through explicit criteria and reasoning.
Prior work explores prompt engineering~\cite{zheng2023judging}, human-like workflows~\cite{zhang2025crowd}, CoT-augmented preference tuning~\cite{mahan2024generative,trivedi2024self}, and RL-induced reasoning behaviors, as in DeepSeek-GRM~\cite{liu2025inference}, RM-R1~\cite{chen2025rm}, J1~\cite{whitehouse2025j1}, and BR-RM~\cite{jiao2025think}. However, elaborate reasoning incurs substantial computational overhead, conflicting with efficiency requirements and potentially inducing overthinking~\cite{yue2025don}. Most closely related to our work, \citet{xu2026ask} route uncertain RM queries to a strong external LLM judge under a fixed budget, while recent work such as CAMEL~\citep{zhu2026camel} also explores adaptive reasoning for reward modeling. Unlike the two-model cascade in \citet{xu2026ask}, F/S-RM integrates Fast and Slow thinking within a single backbone, enabling direct mode switching and prefix reuse without relying on an external judge.

\section{Conclusion}
In this work, we introduce F/S-RM, a unified reward modeling framework that integrates fast preference judgment and slow deliberative reasoning within a single backbone. Inspired by dual-process thinking, F/S-RM first makes an efficient first-token judgment and selectively activates chain-of-thought reasoning only when the initial decision is uncertain. A two-stage training strategy enables the two modes to coexist within the same model, while the dual-confidence mechanism provides adaptive control over their activation. Extensive experiments demonstrate that Fast and Slow thinking are complementary rather than interchangeable, allowing F/S-RM to preserve strong judgment capability while avoiding unnecessary reasoning. More broadly, our results suggest that reward modeling and LLM-as-a-Judge systems need not choose between efficient direct prediction and costly deliberative reasoning; instead, reasoning computation can be allocated adaptively according to the uncertainty of each judgment.

\section{Limitations}
While F/S-RM effectively integrates fast and slow thinking within a unified reward model, several limitations remain. Our experiments focus on Qwen3-4B and Qwen3-8B, and the proposed training and activation mechanisms have not yet been systematically validated at substantially larger scales, where confidence calibration and Fast--Slow interactions may differ. The current implementation also focuses on pairwise preference judgment; extending F/S-RM to pointwise scoring, listwise ranking, and rubric- or criteria-based fine-grained evaluation would broaden its applicability.

The dual-confidence mechanism assumes that uncertainty in Fast judgment is informative for deciding when additional reasoning is needed. However, confidence is not perfectly aligned with correctness, so confidently incorrect predictions may bypass Slow reasoning. More robust uncertainty estimation or learned activation strategies could improve this process.

\section*{Acknowledgments}

This work is supported by Meituan, National Natural Science Foundation of China (NSFC) under the Grant No.~62372113, and Major Project of the National Social Science Foundation of China (NSSFC) under the Grant No.~25\&ZD260. Peng Zhang \& Tun Lu are with the College of Computer Science and Artificial Intelligence, Fudan University. Tun Lu is also affiliated to the Silver-X MOE Philosophy \& Social Sciences Laboratory, Fudan Institute on Aging, and Shanghai Key Laboratory of Data Science, Fudan University.


\FloatBarrier  
\bibliography{custom}

\appendix

\section{Appendix}
\subsection{Training Dataset and Parameter}
\label{sec:Training Dataset}
\begin{table}[h]
\centering

\label{tab:global_stats}
\begin{tabular}{lcc}
\hline
\textbf{Source} & \textbf{Size} & \textbf{Domain} \\
\hline
skywork-single          & 14621  & Chat \\
Code-Preference-Pairs   & 3027  & Reasoning \\
Math-DPO-10K              & 2352 & Reasoning \\
\hline
\end{tabular}
\caption{Global Statistics of our Training Dataset. For the construction of the original full dataset, please refer to Chen et al.\cite{chen2025rm}}
\end{table}

\begin{table}[t]
\centering

\small
\setlength{\tabcolsep}{4pt}
\renewcommand{\arraystretch}{1.1}
\begin{tabular}{@{}c p{3.2cm} p{2.5cm}@{}}
\toprule
\textbf{Category} & \textbf{Parameter} & \textbf{Value} \\
\midrule

\multirow{7}{*}{\textbf{SFT}} 
& Optimization Algorithm & AdamW \\
& Training Epochs & 2 \\
& Batch Size & 32 \\
& Learning Rate 4B & $5 \times 10^{-7}$ \\
& Learning Rate 8B & $1 \times 10^{-6}$ \\
& Precision & bfloat16 \\
& $\lambda$ & 1.0 \\
& lr scheduler type  & cosine \\

\midrule

\multirow{13}{*}{\textbf{RL}} 
& Optimization Algorithm & GRPO \\
& Training Steps & 400 \\
& Global Batch Size & 64 \\
& Micro Batch Size & 16 \\
& Group Size & 8 \\
& Temperature & 1.0 \\
& PPO Clip (Min/Max) & 0.2/0.2 \\
& KL Penalty & 0.001 \\
& Weight Decay & 0.0 \\
& Gradient Clip Norm & 1.0 \\
& Learning Rate & $1 \times 10^{-6}$ \\
& Max Total Tokens & 4096 \\
& Precision & bfloat16 \\

\midrule

\multirow{5}{*}{\textbf{Evaluation}} 
& Backend & vLLM \\
& Temperature & 0.6 \\
& Sampling Repetitions & 8 \\
& Top-P & 0.95 \\
& Top-K & 20 \\
& Max Total Tokens & 8,192 \\
& K of Token Conf & 20 \\

\bottomrule
\end{tabular}
\caption{Training and Inference Configurations.}
\label{tab:training_config}
\end{table}

\subsection{Evaluation Datasets}
\label{sec:Evaluation Datasets}
For evaluation, we select three main benchmark test suites:

\paragraph{RewardBench\cite{lambert2025rewardbench}} As the most widely used evaluation dataset for reward models, model performance on RewardBench is already close to saturation. However, this dataset has been pointed out to contain certain sources of bias. For example, in mathematical prompts, the preferred answer often places the final result inside \verb|\boxed{}|, while the rejected answer instead places the result after \verb|# Answer|\cite{wu2025rewordbench}. This design makes RewardBench more favorable to reward models trained on preference data generated by GPT-4. Despite these drawbacks, during the period of this work, RewardBench remains a strong and popular benchmark, so we continue to adopt it.

\paragraph{RM-Bench\cite{liu2024rm}} RM-Bench is regarded as an effective alternative to RewardBench. It covers task categories similar to RewardBench, such as Chat, Safety, Math, and Code, but is designed with a stronger focus on three aspects: sensitivity to subtle content differences, robustness against stylistic biases, and correlation with the performance of policy models. It mainly addresses the limitations of existing benchmarks in detecting sensitivity to fine-grained content differences and in measuring robustness to stylistic variations (e.g., length and formatting).

\paragraph{JudgeBench\cite{tan2024judgebench}} JudgeBench is a benchmark dataset targeting objective judgment in complex tasks. It primarily mitigates the issue that many existing benchmarks overly rely on subjective preference signals, making it difficult to accurately evaluate a model's ability to distinguish factual correctness in challenging domains such as mathematics, coding, and reasoning. JudgeBench is a highly challenging dataset: even GPT-4o or top-tier reward models achieve only about 64\% accuracy on this benchmark, making it well-suited for testing the intrinsic reasoning ability of reward models.
\subsection{Prompts Used in Our Experiments}

\begin{figure}[th]
\centering
\begin{tcolorbox}[
    enhanced,
    breakable,
    width=\linewidth,
    colback=teal!5,
    colframe=teal!60!black,
    boxrule=0.5pt,
    arc=3pt,
    left=5pt,right=5pt,top=5pt,bottom=5pt,
    title={Adaptive Reasoning Prompt},
    coltitle=white,
    colbacktitle=teal!70!black,
    fonttitle=\bfseries,
]

\small
\textbf{Instructions:} Please act as an impartial judge and evaluate the quality of the responses provided by two AI assistants to the user question shown below. You should choose the assistant that follows the user's instructions and answers the question better. Your evaluation should consider factors such as helpfulness, relevance, accuracy, depth, creativity, and level of detail.

\begin{enumerate}
    \item Begin by making a preliminary judgment. Output \texttt{'A'}, \texttt{'B'}, or \texttt{'tie'} as the first token:
    \begin{itemize}
        \item Output \texttt{'A'} or \texttt{'B'} if you are confident in your immediate judgment. End your response there.
        \item Output \texttt{'tie'} if you are unable to judge immediately and require further reasoning.
    \end{itemize}
    \item If you output \texttt{'tie'}, provide a step-by-step reanalysis comparing the nuances within \texttt{\textless think\textgreater} and \texttt{\textless /think\textgreater} tags.
    \item After your analysis, output your final definitive judgment (\texttt{'A'} or \texttt{'B'}) as the last token.
\end{enumerate}

\vspace{0.3em}
[User Question]\\
\textit{<user question text>}

\vspace{0.3em}
[The Start of Assistant A's Answer]\\
\textit{<Assistant A's response>}

\vspace{0.3em}
[The End of Assistant A's Answer]

\vspace{0.3em}
[The Start of Assistant B's Answer]\\
\textit{<Assistant B's response>}

\vspace{0.3em}
[The End of Assistant B's Answer]

\end{tcolorbox}
\caption{Adaptive Reasoning Prompt}
\label{fig:system_prompt}
\end{figure}

Figure~\ref{fig:system_prompt} presents the adaptive reasoning prompt used in our evaluation framework. The prompt instructs the judge model to first make a fast preliminary judgment by outputting \texttt{'A'}, \texttt{'B'}, or \texttt{'tie'}. If the model outputs \texttt{'tie'}, indicating uncertainty, it switches to slow thinking mode and provides a detailed step-by-step analysis within \texttt{<think>} tags before making a final judgment. This two-stage approach allows the model to adaptively allocate computational resources based on task difficulty, using fast reasoning for clear-cut cases and slow reasoning for more nuanced comparisons.

\subsection{Detailed Ablation on Dual-Confidence Mechanism}
\label{app:dual_confidence_ablation}

\begin{table*}[!t]
\centering

\resizebox{\textwidth}{!}{
\begin{tabular}{ll|ccc|ccc|ccc}
\toprule
\multirow{2}{*}{\textbf{Model}} & \multirow{2}{*}{\textbf{Strategy}} & \multicolumn{3}{c|}{\textbf{RewardBench}} & \multicolumn{3}{c|}{\textbf{RM-Bench}} & \multicolumn{3}{c}{\textbf{JudgeBench}} \\
& & \textbf{Acc} & \textbf{$\Delta$Slow} & \textbf{Token$\downarrow$} & \textbf{Acc} & \textbf{$\Delta$Slow} & \textbf{Token$\downarrow$} & \textbf{Acc} & \textbf{$\Delta$Slow} & \textbf{Token$\downarrow$} \\
\midrule
\multirow{5}{*}{\textbf{Qwen3-4B}} 
& Slow  & 90.50 & - & 0\% & 84.59 & - & 0\% & 69.43 & - & 0\% \\
\cmidrule{2-11}
& \cellcolor{blue!10}Dual-Conf (Hybrid) & \cellcolor{blue!10}\textbf{90.45} & \cellcolor{blue!10}\textbf{-0.05} & \cellcolor{blue!10}31.18\% & \cellcolor{blue!10}\textbf{84.48} & \cellcolor{blue!10}\textbf{-0.11} & \cellcolor{blue!10}19.91\% & \cellcolor{blue!10} 69.29 & \cellcolor{blue!10}-0.14 & \cellcolor{blue!10}6.45\% \\
& Intuition-Conf ($C^I$) & 90.32 & -0.18 & \textbf{56.89\%} & 83.81 & -0.78 & \textbf{40.08\%} & \textbf{69.71} & \textbf{+0.28} & \textbf{29.21\%} \\
& Token-Conf ($C^T$) & 89.55 & -0.95 & 39.84\% & 84.14 & -0.45 & 24.07\% & 69.29 & -0.14 & 7.18\% \\
\cmidrule{2-11}
& \textit{Correlation ($C^I$, $C^T$)} & \multicolumn{3}{c|}{\textit{0.4671}} & \multicolumn{3}{c|}{\textit{0.6368}} & \multicolumn{3}{c}{\textit{0.3724}} \\
\midrule
\multirow{5}{*}{\textbf{Qwen3-8B}} 
& Slow  & 91.94 & - & 0\% & 88.23 & - & 0\% & 72.43 & - & 0\% \\
\cmidrule{2-11}
& \cellcolor{blue!10}Dual-Conf (Hybrid) & \cellcolor{blue!10}\textbf{92.03} & \cellcolor{blue!10}\textbf{+0.09} & \cellcolor{blue!10}39.07\% & \cellcolor{blue!10}\textbf{87.89} & \cellcolor{blue!10}\textbf{-0.34} & \cellcolor{blue!10}22.29\% & \cellcolor{blue!10}73.00 & \cellcolor{blue!10}+0.57 & \cellcolor{blue!10}6.24\% \\
& Intuition-Conf ($C^I$) & 91.56 & -0.38 & \textbf{57.66\%} & 87.20 & -1.03 & 36.53\% & \textbf{73.43} & \textbf{+1.00} & \textbf{28.76\%} \\
& Token-Conf ($C^T$) & 91.81 & -0.13 & 42.85\% & 86.66 & -1.57 & \textbf{41.03\%} & 71.57 & -0.86 & 14.47\% \\
\cmidrule{2-11}
& \textit{Correlation ($C^I$, $C^T$)} & \multicolumn{3}{c|}{\textit{0.6759}} & \multicolumn{3}{c|}{\textit{0.5325}} & \multicolumn{3}{c}{\textit{0.2602}} \\
\bottomrule
\end{tabular}
}
\caption{Ablation study on confidence-based routing strategies across three benchmarks with different correlation levels. \textbf{$\Delta$Slow}: Accuracy difference compared to Slow mode baseline. \textbf{Token$\downarrow$}: Percentage of token savings. \textbf{Correlation}: Pearson correlation coefficient between intuition confidence $C^I$ and token confidence $C^T$. Results are shown for both Qwen3-4B and Qwen3-8B base models, which use dual-confidence thresholds of ($\tau_I=0.641$, $\tau_T=14.224$) and ($\tau_I=0.642$, $\tau_T=15.108$) respectively.}
\label{tab:ablation_strategy}
\end{table*}
Table~\ref{tab:ablation_strategy} compares dual-confidence routing against single-confidence baselines ($C^I$ only and $C^T$ only) across three benchmarks. The dual-confidence mechanism achieves the best accuracy-efficiency trade-off: it maintains performance closest to Slow mode while achieving moderate token savings. Single-confidence strategies show higher token savings but with greater performance degradation. Notably, the effectiveness of each strategy correlates with the confidence signal correlation—dual-confidence performs best when $C^I$ and $C^T$ exhibit moderate correlation (0.47-0.64), as complementary signals provide more robust routing decisions.
\subsection{Routing Reliability and Threshold Sensitivity}
\label{app:routing_reliability}

We further analyze the failure modes of confidence-based routing. Let a sample be \emph{Confident} if it is assigned to the Direct path and \emph{Uncertain} if it is routed to Slow thinking. A missed escalation occurs when Fast is wrong, Slow would be correct, but the sample is classified as Confident and therefore does not trigger Slow reasoning.

\begin{table*}[t]
\centering
\small
\setlength{\tabcolsep}{5pt}
\begin{tabular}{@{}lrrrr@{}}
\toprule
\textbf{Routing outcome}
& \textbf{RM-Bench}
& \textbf{JudgeBench}
& \textbf{RewardBench}
& \textbf{Overall} \\
\midrule
Fast accuracy
& 80.3\% & 65.1\% & 87.5\% & -- \\
Slow accuracy
& 88.2\% & 72.4\% & 91.9\% & -- \\
\addlinespace
Confident $\rightarrow$ Direct
& 2,860 (35.9\%) & 59 (8.4\%) & 2,768 (46.4\%) & 5,687 (38.9\%) \\
\quad Direct-routed Fast errors
& 142 & 4 & 68 & 214 \\
\quad\quad Slow correct
& 46 (0.58\%) & 0 (0.0\%) & 18 (0.30\%) & \textbf{64 (0.44\%)} \\
\quad\quad Slow wrong
& 96 & 4 & 50 & 150 \\
\addlinespace
Uncertain, Fast wrong, Slow correct
& 906 (11.4\%) & 135 (19.3\%) & 418 (7.0\%) & \textbf{1,459 (10.0\%)} \\
\bottomrule
\end{tabular}
\caption{Detailed routing-reliability analysis on Qwen3-8B. ``Slow correct'' under direct-routed Fast errors denotes avoidable missed escalations: Fast was wrong and Slow would have corrected the prediction, but the instance was classified as confident.}
\label{tab:routing_breakdown}
\end{table*}

Among the 1,523 Slow-fixable Fast errors, the router sends 1,459 to Slow reasoning, corresponding to a capture rate of 95.8\%. The remaining 64 missed escalations account for only 0.44\% of all evaluated samples. Nevertheless, these cases demonstrate that confidence and correctness are not perfectly aligned and that confidently wrong Fast predictions remain an inherent failure mode of confidence-based routing.

\subsection{Prefix Reuse Analysis}
\label{app:Prefix Reuse}

When Hybrid inference activates the Slow path, the Slow prompt differs from the Fast prompt only by the appended \texttt{tie} trigger. We reuse the Fast-stage KV cache through the native Hugging Face cache interface, avoiding redundant prefill computation.

\begin{table}[t]
\centering
\small
\setlength{\tabcolsep}{4pt}
\begin{tabular}{@{}lrrrr@{}}
\toprule
\textbf{Dataset} & \textbf{Input} & \textbf{Reused} & \textbf{New} & \textbf{Reuse} \\
\midrule
RewardBench & 599  & 598  & 1 & 99.83\% \\
RM-Bench    & 1,006 & 1,005 & 1 & 99.90\% \\
JudgeBench  & 1,660 & 1,659 & 1 & 99.94\% \\
\bottomrule
\end{tabular}
\caption{Prefix reuse when Hybrid inference enters the Slow path. Input, Reused, and New denote the Slow-path input tokens, reused prefix tokens, and newly appended trigger tokens, respectively.}
\label{tab:prefix_reuse}
\end{table}

Prefix reuse primarily reduces the startup cost of triggered Slow reasoning rather than the decoding cost of the subsequent reasoning chain. For Qwen3-8B, the latency to the first Slow token decreases from 94.3 ms without reuse to 35.5 ms with reuse, corresponding to a 62.4\% reduction. 
In contrast, Table~\ref{tab:hybrid_latency} shows only a small difference in total Hybrid latency with and without reuse, confirming that the overall efficiency gain mainly comes from samples that terminate through the Fast path.

\begin{figure*}[h]
    \centering
    \includegraphics[width=1\linewidth]{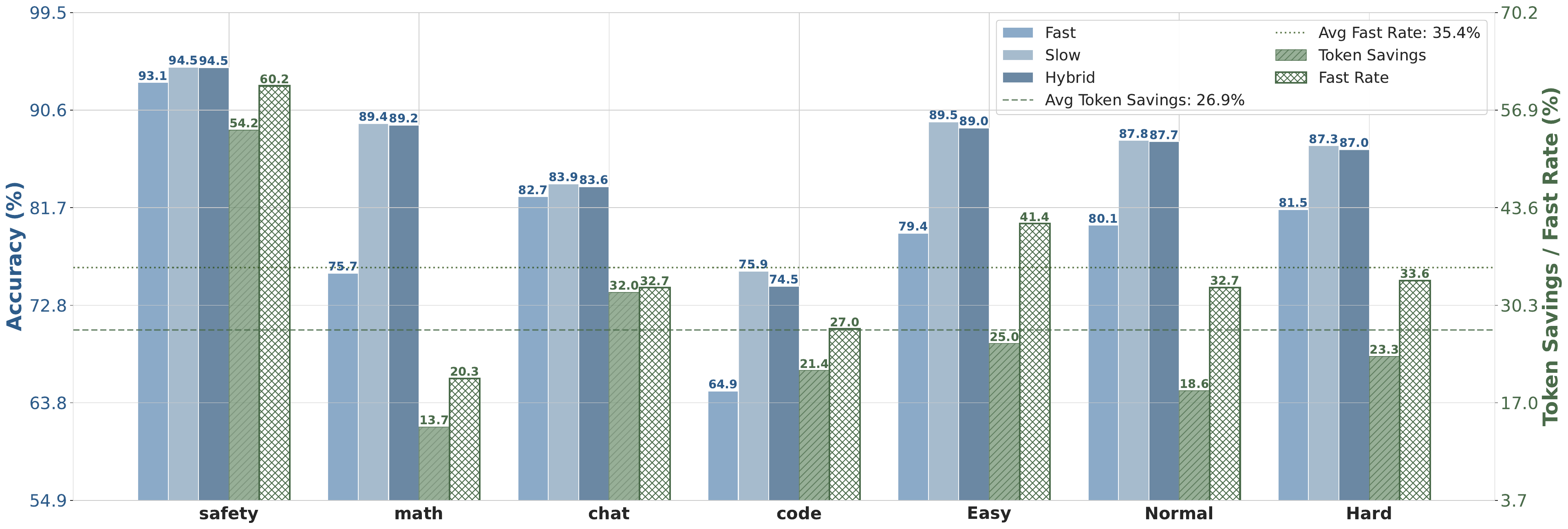}
    \caption{Performance breakdown across domains and difficulty levels on RM-Bench. Left y-axis (blue bars): accuracy under Fast, Slow, and Hybrid modes. Right y-axis (green bars): Fast Rate and Token Saving percentages for Hybrid mode.}
    \label{fig:domain_ana}
\end{figure*}

\subsection{Metric Calculation}
\label{app: Metric}
This appendix defines the two key evaluation metrics used in our study.

\textbf{Fast Rate}

The \textit{Fast Rate} measures the proportion of samples that use only the Fast mode without triggering Slow thinking in the hybrid system:

\begin{equation}
\text{FastRate}_d = \frac{\sum_{s \in S_d} \mathbbm{1}[\text{use\_only\_fast}(s)]}{|S_d|}
\end{equation}

where $S_d$ denotes the sample set in domain $d$, and $\mathbbm{1}[\text{use\_only\_fast}(s)] = 1$ if sample $s$ uses only Fast mode without activating Slow thinking, $0$ otherwise.

\textbf{Token Savings}

The \textit{Token Savings} metric quantifies the reduction in token consumption achieved by Hybrid mode compared to Slow mode:

\begin{equation}
\text{TokenSavings}_d = 1 - \frac{\sum_{s \in S_d} \text{tokens}_{\text{hybrid}}(s)}{\sum_{s \in S_d} \text{tokens}_{\text{slow}}(s)}
\end{equation}

where $\text{tokens}_{\text{hybrid}}(s)$ and $\text{tokens}_{\text{slow}}(s)$ represent the token consumption of sample $s$ in Hybrid and Slow modes, respectively.
\subsection{Declaration of AI Assistance}
The authors used AI tools strictly and exclusively for linguistic refinement, such as grammar correction and improving sentence flow. All core ideas, research methodologies, and key technical innovations presented in this work were conceived and developed independently by the authors without the assistance of AI. The authors have reviewed all AI-suggested stylistic modifications and assume full responsibility for the integrity, originality, and accuracy of the entire manuscript

\subsection{Fine-grained Domain-Specific Comparison}
\label{sec:Domain-Specific Comparison}

To verify that fast and slow thinking are not redundant, we compare their correctness patterns on Qwen3-8B. Table~\ref{tab:complementarity} reports the confusion matrix between fast-only and slow-only decisions. The ``Slow Fixes'' cells indicate cases where slow reasoning corrects fast mistakes, while the ``Fast Corrects'' cells indicate cases where fast judgment succeeds but slow reasoning fails, often due to overthinking or sensitivity to superficial reasoning traces. The resulting oracle ceilings are substantially higher than the current hybrid performance, suggesting that the main remaining opportunity lies in better routing rather than simply increasing CoT length.
The sizeable Fast Corrects population shows that slow thinking is not uniformly superior: it can overthink easy or bias-sensitive cases. Conversely, the Slow Fixes population confirms the value of deliberate reasoning on ambiguous examples. This asymmetry explains why adaptive routing can outperform slow-only inference on JudgeBench and why future gains should focus on richer uncertainty signals for mode selection.

Furthermore, we provide a comprehensive analysis of our hybrid system's performance, we conduct fine-grained comparisons across different domains and difficulty levels. This section presents domain-specific performance breakdowns for three benchmark datasets: Rewardbench \ref{tab:rm_comparison}, RM-bench \ref{tab:rm-b_results}, and Judgebench \ref{tab:judgebench}.

\begin{table*}[h]
\centering
\small
\begin{tabular}{lccccc}
\toprule
Models & Chat & Chat\_Hard & Safety & Reasoning & Overall \\
\midrule
\multicolumn{6}{l}{\textbf{SRMs}} \\
\midrule
Eurus-RM-7b & 98.0 & 65.6 & 81.4 & 86.3 & 82.8 \\
Internlm2-7b-reward & \textbf{99.2} & 69.5 & 87.2 & 94.5 & 87.6 \\
SteerLM-RM-70B & 91.3 & 80.3 & 92.8 & 90.6 & 88.8 \\
Cohere-0514 & 96.4 & 71.3 & 92.3 & 97.7 & 89.4 \\
Internlm2-20b-reward & 98.9 & 76.5 & 89.5 & 95.8 & 90.2 \\
ArmoRM-Llama3-8B-v0.1 & 96.9 & 76.8 & 90.5 & 97.3 & 90.4 \\
Nemotron-4-340B-Reward & 95.8 & 87.1 & 91.5 & 93.6 & 92.0 \\
Skywork-Reward-Llama-3.1-8B$^\dagger$ & 95.8 & 87.3 & 90.8 & 96.2 & 92.5 \\
Skywork-Reward-Gemma-2-27B$^\dagger$ & 95.8 & \textbf{91.4} & 91.9 & 96.1 & 93.8 \\
INF-ORM-Llama3.1-70B & 96.6 & 91.0 & \textbf{93.6} & \textbf{99.1} & \textbf{95.1} \\
\midrule
\multicolumn{6}{l}{\textbf{GRMs}} \\
\midrule
Llama3.1-8B-Instruct & 85.5 & 48.5 & 75.6 & 72.1 & 70.4 \\
Prometheus-8$\times$7B-v2 & 93.0 & 47.1 & 80.5 & 77.4 & 74.5 \\
Llama3.1-70B-Instruct & 97.2 & 70.2 & 82.8 & 86.0 & 84.0 \\
Llama3.1-405B-Instruct & 97.2 & 74.6 & 77.6 & 87.1 & 84.1 \\
Claude-3.5-sonnet-20240620 & 96.4 & 74.0 & 81.6 & 84.7 & 84.2 \\
GPT-4o-0806 & 96.1 & 76.1 & 86.6 & 88.1 & 86.7 \\
Gemini-1.5-pro & 92.3 & 80.6 & 87.9 & 92.0 & 88.2 \\
SFR-LLaMa-3.1-70B-Judge-r & 96.9 & 84.8 & 91.6 & 97.6 & 92.7 \\
Skywork-Critic-Llama-3.1-70B$^\dagger$ & 96.6 & 87.9 & 93.1 & 95.5 & 93.3 \\
JudgeLRM & 92.9 & 56.4 & 78.2 & 73.6 & 75.2 \\
SynRM & 38.0 & 82.5 & 74.1 & 87.1 & 70.4 \\
RM-R1-DeepSeek-Distilled-Qwen-7B & 88.9 & 66.2 & 78.4 & 87.0 & 80.1 \\
CLoud & 97.0 & 58.0 & 84.0 & 92.0 & 82.8 \\
DeepSeek-GRM-16B & 90.8 & 74.3 & 84.7 & 81.8 & 82.9 \\
DeepSeek-GRM-27B-RFT & 94.7 & 77.2 & 87.0 & 79.2 & 84.5 \\
RM-R1-Qwen-Instruct-7B & 94.1 & 74.6 & 85.2 & 86.7 & 85.2 \\
DeepSeek-GRM-27B & 94.1 & 78.3 & 88.0 & 83.8 & 86.0 \\
DeepSeek-PairRM-27B & 95.5 & 86.8 & 52.3 & 92.0 & 87.1 \\
RM-R1-Qwen-Instruct-14B & 93.6 & 80.5 & 86.9 & 92.0 & 88.2 \\
RM-R1-DeepSeek-Distilled-Qwen-14B & 91.3 & 79.4 & 89.3 & 95.5 & 88.9 \\
Self-taught-evaluator-llama3.1-70B & 96.9 & 85.1 & 89.6 & 88.4 & 90.0 \\
RM-R1-DeepSeek-Distilled-Qwen-32B & 95.3 & 80.3 & 91.1 & 96.8 & 90.9 \\
RM-R1-Qwen-Instruct-32B & 95.3 & 83.1 & 91.9 & 95.2 & 91.4 \\
BR-RM-Qwen-8B & 95.8 & 80.1 & 90.4 & 97.5 & 91.0 \\
BR-RM-Qwen-14B & 97.0 & 82.4 & 90.0 & 98.8 & 92.1 \\
Uncertainty-Router-8B & 98.3 & 81.2 & 90.0 & 97.0 & 91.6  \\
\midrule
\multicolumn{6}{@{}l@{}}{\textbf{Our F/S-RM}} \\
\midrule
F/S-RM-Qwen3-4B-Hybrid & 93.7 & 80.2 & 88.9 & 93.7 & 90.5 \\
\rowcolor{cyan!15}
\quad Fast Rate & \textcolor{blue!70!black}{57.8\%} & \textcolor{blue!70!black}{53.8\%} & \textcolor{blue!70!black}{78.9\%} & \textcolor{blue!70!black}{17.9\%} & \textcolor{blue!70!black}{34.7} \\
\rowcolor{cyan!15}
\quad $\downarrow$ Token Savings & \textcolor{green!70!black}{53.5\%} & \textcolor{green!70!black}{49.1\%} & \textcolor{green!70!black}{76.1\%} & \textcolor{green!70!black}{11.8\%} & \textcolor{green!70!black}{31.2\%} \\
F/S-RM-Qwen3-4B-Fast & 90.9 & 78.2 & 86.4 & 86.8 & 85.9 \\
F/S-RM-Qwen3-4B-Slow & 93.9 & 80.3 & 88.9 & 93.8 & 90.5 \\
\midrule
F/S-RM-Qwen3-8B-Hybrid & 94.3 & 85.2 & 89.4 & 95.0  & 92.0 \\
\rowcolor{cyan!15}
\quad Fast Rate & \textcolor{blue!70!black}{38.1\%} & \textcolor{blue!70!black}{38.4\%} & \textcolor{blue!70!black}{53.1\%} & \textcolor{blue!70!black}{47.5\%} & \textcolor{blue!70!black}{44.3\%} \\
\rowcolor{cyan!15}
\quad $\downarrow$ Token Savings & \textcolor{green!70!black}{36.5\%} & \textcolor{green!70!black}{34.2\%} & \textcolor{green!70!black}{38.8\%} & \textcolor{green!70!black}{46.1\%} & \textcolor{green!70!black}{39.1\%} \\
F/S-RM-Qwen3-8B-Fast & 88.1 & 84.2 & 86.3 & 89.1  & 87.5 \\
F/S-RM-Qwen3-8B-Slow & 94.4 & 85.0 & 89.5 & 94.8  & 91.9 \\
\bottomrule
\end{tabular}
\caption{The full results of tested reward models on \textbf{RewardBench}. Chat, Chat\_hard, Safety, and Reasoning show the model’s Average accuracy on each domain. Bold numbers indicate the best performance.}
\label{tab:rm_comparison}
\end{table*}




\begin{table*}[t]
\centering
\small
\setlength{\tabcolsep}{3pt}
\renewcommand{\arraystretch}{1.05}
\begin{tabular}{@{}lcccccccc@{}}
\toprule
\textbf{Model} & \textbf{Chat} & \textbf{Math} & \textbf{Code} & \textbf{Safety} & \textbf{Easy} & \textbf{Normal} & \textbf{Hard} & \textbf{Avg} \\
\midrule
\multicolumn{9}{@{}l@{}}{\textbf{SRMs}} \\
\midrule
steerlm-70b & 56.4 & 53.0 & 49.3 & 51.2 & 48.3 & 54.9 & 54.3 & 52.5 \\
tulu-v2.5-70b-preference-mix-rm & 58.2 & 51.4 & 55.5 & 87.1 & 72.8 & 65.6 & 50.7 & 63.0 \\
Mistral-7B-instruct-Unified-Feedback & 56.5 & 58.0 & 51.7 & 86.8 & 87.1 & 67.3 & 35.3 & 63.2 \\
RM-Mistral-7B & 57.4 & 57.0 & 52.7 & 87.2 & 88.6 & 67.1 & 34.9 & 63.5 \\
Eurus-RM-7b & 59.9 & 60.2 & 56.9 & 86.5 & 87.2 & 70.2 & 40.2 & 65.9 \\
internlm2-7b-reward & 61.7 & 71.4 & 49.7 & 85.5 & 85.4 & 70.7 & 45.1 & 67.1 \\
Skywork-Reward-Gemma-2-27B & 69.5 & 54.7 & 58.2 & 91.9 & 78.0 & 69.2 & 54.9 & 67.3 \\
ArmorRM-Llama3-8B-v0.1 & 67.8 & 57.5 & 53.1 & 92.4 & 82.2 & 71.0 & 49.8 & 67.7 \\
GRM-llama3-8B-sftreg & 62.7 & 62.5 & 57.8 & 90.0 & 83.5 & 72.7 & 48.6 & 68.2 \\
internlm2-20b-reward & 63.1 & 66.8 & 56.7 & 86.5 & 82.6 & 71.6 & 50.7 & 68.3 \\
llama-3-OffsetBias-RM-8B & 71.3 & 61.9 & 53.2 & 89.6 & 84.6 & 72.2 & 50.2 & 69.0 \\
Nemotron-340B-Reward & 71.2 & 59.8 & 59.4 & 87.5 & 81.0 & 71.4 & 56.1 & 69.5 \\
URM-LLaMA-3.1-8B & 71.2 & 61.8 & 59.1 & 93.1 & 84.0 & 73.2 & 53.0 & 70.0 \\
Skywork-Reward-Llama-3.1-8B & 69.5 & 60.6 & 54.5 & 95.7 & 89.0 & 74.7 & 46.6 & 70.1 \\
INF-ORM-Llama3.1-70B & 66.3 & 65.6 & 56.8 & 94.8 & 91.8 & 76.1 & 44.8 & 70.9 \\
\midrule
\multicolumn{9}{@{}l@{}}{\textbf{GRMs}} \\
\midrule
tulu-v2.5-dpo-13b-chatbot-arena-2023 & 64.9 & 52.3 & 50.5 & 62.3 & 82.8 & 60.2 & 29.5 & 57.5 \\
tulu-v2.5-dpo-13b-nectar-60k & 56.3 & 52.4 & 52.6 & 73.8 & 86.7 & 64.3 & 25.4 & 58.8 \\
stablelm-2-12b-chat & 67.2 & 54.9 & 51.6 & 65.2 & 69.1 & 63.5 & 46.6 & 59.7 \\
tulu-v2.5-dpo-13b-stackexchange-60k & 66.4 & 49.9 & 54.2 & 69.0 & 79.5 & 63.0 & 37.2 & 59.9 \\
Nous-Hermes-2-Mistral-7B-DPO & 58.8 & 55.6 & 51.3 & 73.9 & 69.5 & 61.1 & 49.1 & 59.9 \\
Claude-3-5-sonnet-20240620 & 62.5 & 62.6 & 54.4 & 64.4 & 73.8 & 63.4 & 45.9 & 61.0 \\
tulu-v2.5-dpo-13b-hh-rlhf-60k & 68.4 & 51.1 & 52.3 & 76.5 & 53.6 & 63.0 & 69.6 & 62.1 \\
tulu-2-dpo-13b & 66.4 & 51.4 & 57.8 & 85.4 & 86.9 & 66.7 & 37.7 & 63.8 \\
SOLAR-10.7B-Instruct-v1.0 & 78.6 & 52.3 & 49.6 & 78.9 & 57.5 & 67.6 & 69.4 & 64.8 \\
Llama3.1-70B-Instruct & 64.3 & 67.3 & 47.3 & 83.0 & 74.7 & 67.8 & 54.1 & 65.5 \\
Skywork-Critic-Llama-3.1-70B & 71.4 & 64.0 & 56.8 & 94.8 & 85.6 & 73.7 & 56.5 & 71.9 \\
GPT-4o-0806 & 67.2 & 67.5 & 63.6 & 91.7 & 83.4 & 75.6 & 58.7 & 72.5 \\
Gemini-1.5-pro & 71.6 & 73.9 & 63.7 & 91.3 & 83.1 & 77.6 & 64.7 & 75.2 \\
DeepSeek-GRM-27B & 72.4 & 59.2 & 68.0 & 91.1 & 84.6 & 76.5 & 57.0 & 72.7 \\
JudgeLRM & 59.9 & 59.9 & 51.9 & 87.3 & 73.2 & 66.2 & 54.8 & 64.7 \\
RM-R1-Qwen-INSTRUCT-7B & 66.6 & 67.0 & 54.6 & 92.6 & 79.2 & 71.7 & 59.7 & 70.2 \\
Self-taught-evaluator-llama3.1-70B & 73.4 & 65.7 & 56.3 & 90.4 & 80.2 & 74.5 & 59.7 & 71.5 \\
RM-R1-DeepSeek-DISTILLED-Qwen-7B & 64.0 & 83.9 & 56.2 & 85.3 & 75.9 & 73.1 & 68.1 & 72.4 \\
RM-R1-Qwen-INSTRUCT-14B & 75.6 & 75.4 & 60.6 & 93.6 & 82.6 & 77.5 & 68.8 & 76.1 \\
RM-R1-Qwen-INSTRUCT-32B & 75.3 & 80.2 & 66.8 & 93.9 & 86.3 & 80.5 & 70.4 & 79.1 \\
RM-R1-DeepSeek-DISTILLED-Qwen-32B & 71.8 & 90.5 & 69.5 & 94.1 & 86.2 & 83.6 & 74.4 & 81.5 \\
EvalPlanner-Llama-3.3-70B-Instruct & -- & -- & -- & -- & 81.1 & 80.8 & 84.3 & 82.1 \\
RM-R1-DeepSeek-DISTILLED-Qwen-14B & 74.2 & 91.8 & 74.1 & \textbf{95.4} & 89.5 & 85.4 & 76.7 & 83.9 \\
RRM-32B & 73.7 & 91.9 & 75.0 & 95.3 & 90.7 & 85.3 & 76.3 & 84.0 \\
BR-RM-Qwen-8B & 76.4 & \textbf{94.1} & 77.0 & 92.7 & 91.7 & 87.3 & 76.1 & 85.0 \\
BR-RM-Qwen-14B & 77.3 & 92.6 & \textbf{79.8} & 93.7 & 92.0 & \textbf{88.1} & 77.6 & 86.1 \\
Uncertainty-Router-8B & 73.2 & 83.5 & 78.3 & 92.7 & \textbf{93.6} & 86.9 & 65.3 & 81.9 \\
\midrule
\multicolumn{9}{@{}l@{}}{\textbf{Our F/S-RM}} \\
\midrule
F/S-RM-Qwen3-4B-Hybrid & 76.5 & 87.2 & 65.9 & 93.2 & 84.7 & 83.9 & 84.9 & 84.5 \\
\rowcolor{cyan!15}
\quad Fast Rate & \textcolor{blue!70!black}{39.0\%} & \textcolor{blue!70!black}{7.4\%} & \textcolor{blue!70!black}{1.2\%} & \textcolor{blue!70!black}{83.4\%} & \textcolor{blue!70!black}{38.5\%} & \textcolor{blue!70!black}{33.9\%} & \textcolor{blue!70!black}{31.7\%} & \textcolor{blue!70!black}{43.3\%} \\
\rowcolor{cyan!15}
\quad $\downarrow$ Token Savings & \textcolor{green!70!black}{34.0\%} & \textcolor{green!70!black}{4.7\%} & \textcolor{green!70!black}{0.9\%} & \textcolor{green!70!black}{80.8\%} & \textcolor{green!70!black}{20.3\%} & \textcolor{green!70!black}{20.5\%} & \textcolor{green!70!black}{19.0\%} & \textcolor{green!70!black}{19.9\%} \\
F/S-RM-Qwen3-4B-Fast & 75.8 & 73.1 & 61.8 & 92.1 & 77.4 & 77.1 & 78.8 & 77.7 \\
F/S-RM-Qwen3-4B-Slow & 76.6 & 87.2 & 65.9 & 93.4 & 85.2 & 84.0 & 84.5 & 84.6 \\
\midrule
F/S-RM-Qwen3-8B-Hybrid & 83.6 & 89.2 & 74.5 & 94.5 & 89.0 & 87.7 & 87.0 & 87.9 \\
\rowcolor{cyan!15}
\quad Fast Rate & \textcolor{blue!70!black}{32.7\%} & \textcolor{blue!70!black}{20.3\%} & \textcolor{blue!70!black}{27.1\%} & \textcolor{blue!70!black}{60.2\%} & \textcolor{blue!70!black}{41.4\%} & \textcolor{blue!70!black}{32.7\%} & \textcolor{blue!70!black}{33.7\%} & \textcolor{blue!70!black}{35.4\%} \\
\rowcolor{cyan!15}
\quad $\downarrow$ Token Savings & \textcolor{green!70!black}{32.1\%} & \textcolor{green!70!black}{13.7\%} & \textcolor{green!70!black}{21.4\%} & \textcolor{green!70!black}{54.2\%} & \textcolor{green!70!black}{25.1\%} & \textcolor{green!70!black}{18.6\%} & \textcolor{green!70!black}{23.3\%} & \textcolor{green!70!black}{22.3\%} \\
F/S-RM-Qwen3-8B-Fast & 82.7 & 75.7 & 64.9 & 93.1 & 79.4 & 80.1 & 81.5 & 80.3 \\
F/S-RM-Qwen3-8B-Slow & \textbf{83.9} & 89.4 & 75.8 & 94.5 & 89.5 & 87.8 & \textbf{87.4} & \textbf{88.2} \\
\bottomrule
\end{tabular}
\caption{The full results of tested reward models on \textbf{RM-Bench}. Chat, Math, Code, and Safety show the model’s Average accuracy on each domain. Easy, Normal, Hard show the model’s Accuracy on each difficulty level across all domains. Bold numbers indicate the best performance.}
\label{tab:rm-b_results}
\end{table*}

\begin{table*}[t]
\centering
\small
\setlength{\tabcolsep}{4pt}
\renewcommand{\arraystretch}{1.1}
\begin{tabular}{@{}lccccc@{}}
\toprule
\textbf{Model} & \textbf{Knowledge} & \textbf{Reasoning} & \textbf{Math} & \textbf{Coding} & \textbf{Overall} \\
\midrule
\multicolumn{6}{@{}l@{}}{\textbf{Prompted Judges}} \\
\midrule
Vanilla (GPT-4o) & 44.16 & 47.96 & 66.07 & 61.90 & 50.86 \\
Arena-Hard Judge (GPT-4o) & 50.65 & 54.08 & 75.00 & 59.52 & 56.57 \\
VertexAI Evaluation (Gemini-1.5-pro) & 45.45 & 44.90 & 53.57 & 28.57 & 44.57 \\
\midrule
\multicolumn{6}{@{}l@{}}{\textbf{Fine-tuned Judges}} \\
\midrule
PandaLM & 9.09 & 21.43 & 7.14 & 16.67 & 13.14 \\
Prometheus2-7b & 38.31 & 25.51 & 35.71 & 42.86 & 34.86 \\
Prometheus2-8x7b & 41.56 & 39.80 & 50.00 & 23.81 & 40.29 \\
Prometheus2-bgb-8x7b & 45.45 & 30.61 & 46.43 & 28.57 & 39.43 \\
JudgeLM-7B & 23.38 & 29.59 & 32.14 & 11.90 & 25.14 \\
JudgeLM-13B & 26.62 & 29.59 & 28.57 & 19.05 & 26.86 \\
JudgeLM-33B & 32.47 & 48.98 & 33.93 & 19.05 & 35.71 \\
AutoJ & 40.26 & 29.59 & 44.64 & 28.57 & 36.57 \\
Skywork-LLaMA-3.1B-8B & 51.30 & 54.08 & 73.21 & 33.33 & 53.43 \\
Skywork-LLaMA-3.1B-70B & 55.84 & 55.10 & 73.21 & 47.62 & 57.43 \\
\midrule
\multicolumn{6}{@{}l@{}}{\textbf{Proprietary Models}} \\
\midrule
GPT-4o & 50.6 & 54.1 & 75.0 & 59.5 & 59.8 \\
Claude-3.5-Sonnet & 62.3 & 66.3 & 66.1 & 64.3 & 64.8 \\
DeepSeek-R1 & 59.1 & 82.7 & 80.4 & 92.9 & 78.8 \\
o1-preview & 66.2 & 79.6 & 85.7 & 85.7 & 79.3 \\
o3-mini & 58.4 & 62.2 & 82.1 & 78.6 & 70.3 \\
o3-mini (low) & 63.0 & 69.4 & 83.4 & 83.3 & 74.8 \\
o3-mini (medium) & 62.3 & 86.7 & 85.7 & 92.9 & 81.9 \\
o3-mini (high) & \textbf{67.5} & \textbf{89.8} & \textbf{87.5} & \textbf{100} & \textbf{86.2} \\
\midrule
\multicolumn{6}{@{}l@{}}{\textbf{Our F/S-RM}} \\
\midrule
F/S-RM-Qwen3-4B-Hybrid & 57.7 & 77.0 & 82.1 & 70.2 & 69.3 \\
\rowcolor{cyan!15}
\quad Fast Rate & \textcolor{blue!70!black}{8.7\%} & \textcolor{blue!70!black}{13.8\%} & \textcolor{blue!70!black}{9.0\%} & \textcolor{blue!70!black}{1.2\%} & \textcolor{blue!70!black}{9.3\%} \\
\rowcolor{cyan!15}
\quad $\downarrow$ Token Savings & \textcolor{green!70!black}{5.8\%} & \textcolor{green!70!black}{10.1\%} & \textcolor{green!70!black}{5.4\%} & \textcolor{green!70!black}{0.5\%} & \textcolor{green!70!black}{6.5\%} \\
F/S-RM-Qwen3-4B-Fast & 63.6 & 66.8 & 75.4 & 65.5 & 67.0 \\
F/S-RM-Qwen3-4B-Slow & 58.0 & 77.0 & 82.1 & 70.2 & 69.4 \\
\midrule
F/S-RM-Qwen3-8B-Hybrid & 62.9 & 78.1 & 82.1 & 81.0 & 73.0 \\
\rowcolor{cyan!15}
\quad Fast Rate & \textcolor{blue!70!black}{2.8\%} & \textcolor{blue!70!black}{5.6\%} & \textcolor{blue!70!black}{11.9\%} & \textcolor{blue!70!black}{28.6\%} & \textcolor{blue!70!black}{9.7\%} \\
\rowcolor{cyan!15}
\quad $\downarrow$ Token Savings & \textcolor{green!70!black}{1.7\%} & \textcolor{green!70!black}{3.9\%} & \textcolor{green!70!black}{11.1\%} & \textcolor{green!70!black}{22.5\%} & \textcolor{green!70!black}{6.3\%} \\
F/S-RM-Qwen3-8B-Fast & 62.6 & 58.6 & 78.4 & 67.9  & 65.1 \\
F/S-RM-Qwen3-8B-Slow & 61.9 & 77.6 & 82.1 & 81.0  & 72.4 \\
\bottomrule
\end{tabular}
\caption{The full results of tested reward models on \textbf{JudgeBench}. Knowledge, Reasoning, Math, and Coding show the model’s Average accuracy on each domain. Bold numbers indicate the best performance.}
\label{tab:judgebench}
\end{table*}

\end{document}